\documentclass{article}
\usepackage{arxiv}

\usepackage[utf8]{inputenc} % allow utf-8 input
\usepackage[T1]{fontenc}    % use 8-bit T1 fonts
\usepackage{hyperref}       % hyperlinks
\usepackage{url}            % simple URL typesetting
\usepackage{booktabs}       % professional-quality tables
\usepackage{microtype}      % microtypography
\usepackage{lipsum}	    	% Can be removed after putting your text content
\usepackage{graphicx}
\usepackage{doi}
\usepackage[section]{placeins}
\usepackage{mathtools}
\usepackage{amsmath}
\usepackage{amssymb}
\usepackage[ruled,vlined]{algorithm2e}
\title{\textbf{A deep learning energy method for hyperelasticity and viscoelasticity}}

\author{{\hspace{1mm}Diab W. Abueidda}\thanks{abueidd2@illinois.edu} \\
	National Center for Supercomputing Applications\\
	Department of Mechanical Science and Engineering\\
	University of Illinois at Urbana-Champaign\\
	\And
	{\hspace{1mm}Seid Koric}\\
	National Center for Supercomputing Applications\\
	Department of Mechanical Science and Engineering\\
	University of Illinois at Urbana-Champaign\\
	\And
	{\hspace{1mm}Rashid Abu Al-Rub}\\
	Advanced Digital \& Additive Manufacturing Center\\
	Khalifa University of Science and Technology\\
	\And
	{\hspace{1mm}Corey M. Parrott}\\
	Department of Aerospace Engineering\\
	University of Illinois at Urbana-Champaign\\
	\And
	{\hspace{1mm}Kai A. James}\\
	Department of Aerospace Engineering\\
	University of Illinois at Urbana-Champaign\\
	\And
	{\hspace{1mm}Nahil A. Sobh}\\
	National Center for Supercomputing Applications\\
	University of Illinois at Urbana-Champaign\\
}

\renewcommand{\shorttitle}{DEM for hyperelasticity \& viscoelasticity}

\begin{document}

\maketitle

\begin{abstract}
The potential energy formulation and deep learning are merged to solve partial differential equations governing the deformation in hyperelastic and viscoelastic materials. The presented deep energy method (DEM) is self-contained and meshfree. It can accurately capture the three-dimensional (3D) mechanical response without requiring any time-consuming training data generation by classical numerical methods such as the finite element method. Once the model is appropriately trained, the response can be attained almost instantly at any point in the physical domain, given its spatial coordinates. Therefore, the deep energy method is potentially a promising standalone method for solving partial differential equations describing the mechanical deformation of materials or structural systems and other physical phenomena. 
\end{abstract}


\section{Introduction}

Machine learning (ML) has recently been proven effective in many fields such as image and speech recognition, medical diagnoses, autopilot in automotive scenarios, financial services, and many other engineering and medical applications \cite{lim2016speech, thorat2019self, DEBRUIJNE201694}. Computational solid mechanics is no exception. Many researchers have developed data-driven models to capture physical responses \cite{mozaffar2019deep, abueidda2020deep, spear2018data, abueidda2019prediction, sadat2020machine, koric2021deep}. Additionally, data-driven models have been developed to obtain near-optimal topologies for metamaterials and structures, where 2D and 3D domains, linear and nonlinear constraints, and material and geometric nonlinearities have been considered \cite{abueidda2020topology, zheng2020generating, wang2020deep, KOLLMANN2020109098, lin2018investigation}. However, usually one needs a large number of data points to accurately capture intricate relationships between the input and output, making the data generation the bottleneck step in most cases. Furthermore, deep neural networks (DNNs) have been utilized to discover governing equations and material laws using existing computational and/or experimental data \cite{zhao2020discovery, yang2020learning, flaschel2020unsupervised}.

According to the universal approximation theorem, the multilayer feedforward neural networks with an arbitrary nonconstant and bounded activation function and as few as a single hidden layer can serve, with arbitrary accuracy, as universal approximators \cite{cybenko1989approximation, hornik1991approximation}. Given that the activation function is bounded, nonconstant, and continuous, then continuous mappings are uniformly learned over compact sets of inputs. Nevertheless, the theorem neither provides conclusions about the training process, nor the number of neurons required in a hidden layer to obtain a specific accuracy, nor whether the estimation of the network's parameters is even feasible. Several successful trials have been reported for using DNNs to solve partial differential equations (PDEs), which are imperative for describing the physical laws governing all types of phenomena around us. Although solving differential equations using neural networks is not a new topic \cite{meade1994numerical, lagaris1998artificial}, the recent successes are because of: 1) Recent advances in machine learning packages such as PyTorch \cite{NEURIPS2019_9015} and TensorFlow \cite{tensorflow2015-whitepaper} along with advances in CPUs and GPUs, 2) the use of automatic differentiation to accurately find the gradients of functions, and 3) the experience in choosing the networks' architectures that can capture governing equations while satisfying initial and boundary conditions. 

Recently, a number of researchers have proposed and developed frameworks to solve PDEs using DNNs, with a relatively small number of data points or even without the need for any data, by incorporating physical laws into the loss function being minimized \cite{raissi2019physics, abueidda2020meshless, guo2020analysis, he2020physics, haghighat2020deep, guo2020solving, lin2020deep, mahmoudabadbozchelou2021rheology}. The approach used in these papers sometimes is referred to as the deep collocation method (DCM), as collocation points are sampled from the space of interest. Then, the DNNs attempt to find the weights and biases that best satisfy governing equations as well as initial and boundary conditions at the sampled collocation points. This approach is based on the strong form; hence, one usually needs to find second-order derivatives computationally, and the zero and nonzero natural boundary conditions should be explicitly accounted for in the loss function definition.

Weinan et al. \cite{weinan2018deep} proposed a different approach to solve Poisson's equation and eigenvalue problems. In this approach, deep learning and the Ritz method are merged to solve variational problems. Similar to the work of Weinan et al. \cite{weinan2018deep}, Nguyen-Thanh et al. \cite{nguyen2020deep} and Samaniego et al. \cite{samaniego2020energy} proposed a deep energy method (DEM) that utilizes potential energy to solve PDEs appearing in the field of computational solid mechanics. Nonetheless, not all governing equations can be rendered as an energy minimization problem. In the works of Weinan et al. \cite{weinan2018deep}, Nguyen-Thanh et al. \cite{nguyen2020deep}, and Samaniego et al. \cite{samaniego2020energy}, the potential energy is used to define loss functions, where only the first-order derivative is required for a second-order PDE. The advantage of this technique is that the solution procedure only requires first-order automated differentiation rather than second-order as needed for the physics-informed approach based on the strong form. This has the potential to result in faster convergence and greater accuracy. On the other hand, the approach necessitates a successful integration over the domain spanned by the collocation points.  

This study extends the DEM method to solve 3-dimensional partial differential equations for materials obeying a couple of different constitutive models using deep neural networks (DNNs). In this meshfree approach, the DNN predicts the displacement field that satisfies the partial differential equations and the boundary conditions without using any labeled data. Since the training of any ML model is an optimization problem in which the loss function is minimized, we define the loss function using the potential energy. The loss function is minimized using the Adam \cite{kingma2014adam} and quasi-Newton L-BFGS \cite{liu1989limited} optimizers. The DEM is used to find the mechanical response of a hyperelastic model for rubber elastic materials \cite{lopez2010new}. Additionally, the DEM is used to solve a viscoelastic model, inspired by the standard linear solid (SLS) model, by defining two potentials and casting the problem into a machine learning one. To the best of our knowledge, this is the first time that both the hyperelastic model proposed by Lopez-Pamies \cite{lopez2010new} and the SLS viscoelastic model are solved by merging the potential energy and deep learning. This meshfree approach is straightforward to implement; it does not require solving a linear system of equations and assembling the tangent modulus, which are significant steps in most computational methods, such as finite difference and finite element methods, and can be computationally expensive.

The outline of the paper is as follows: Section \ref{method} presents the details of the approach, where a general problem setup is discussed. In Sections \ref{hyper} and \ref{visco}, the DEM approach is used to solve three-dimensional (3D) examples involving hyperelastic and linear viscoelastic constitutive models, respectively. We conclude the paper in Section \ref{conclu} by highlighting the significant results and stating possible future work directions.
\section{Method} \label{method}

The finite element method (FEM) is commonly used to solve problems with material and/or geometric nonlinearities. When an implicit finite element method is used along with an iterative scheme (e.g., Newton-Raphson), the residual vector and tangent matrix are assembled and then used to solve the corresponding linear system of equations, using a direct or iterative solver, to find the vector of unknowns in each iteration. On the other hand, explicit nonlinear finite element methods do not simultaneously solve a linear system; nevertheless, they are often bounded by conditional stability, requiring small time increments. Furthermore, when explicit FEM is used for quasi-static simulations, one needs to ensure that the inertial effects are insignificant \cite{koric2009explicit}. This paper employs deep learning to determine the displacement field, where the displacement field obtained from the DNN is used to compute stresses, strains, and other variables required to satisfy the minimization of potential energy. Using a meshfree approach, the loss function is minimized within a deep learning framework such as PyTorch \cite{NEURIPS2019_9015} and TensorFlow \cite{tensorflow2015-whitepaper}. Subsequently, we cover a brief introduction to deep learning and then discuss the proposed framework.

\subsection{Deep feedforward neural networks}\label{NN}

\begin{figure}[!htb]
    \centering
    \includegraphics[width=0.65\textwidth]{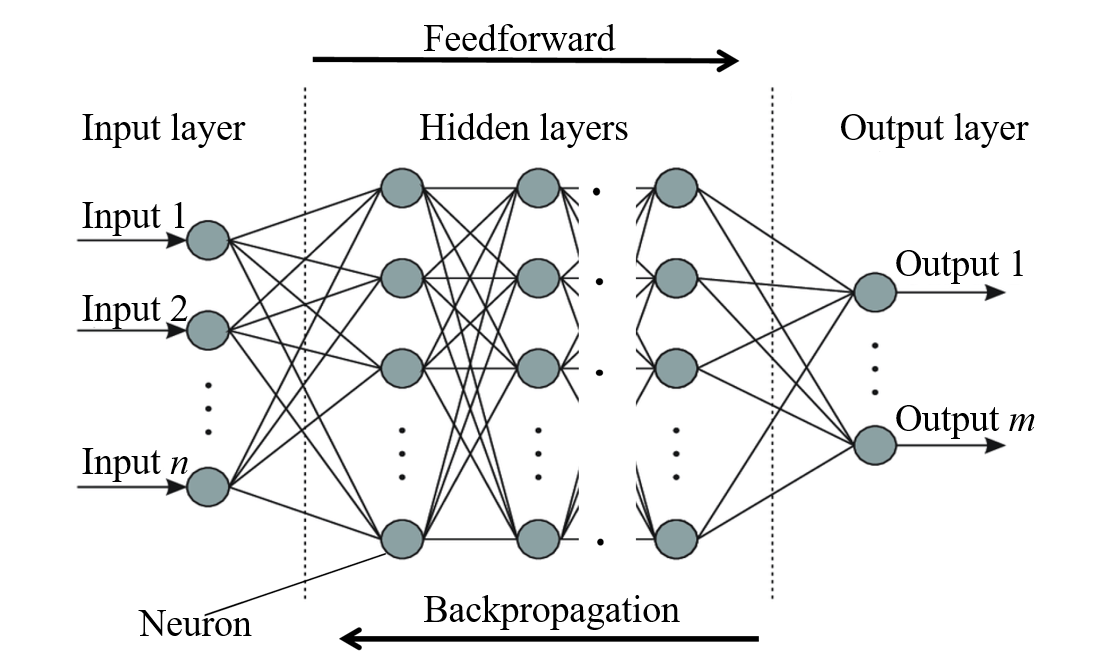}
    \caption{Fully connected (dense) artificial neural network.}
    \label{DANN}
\end{figure}

Deep learning is a subset of machine learning inspired by the configuration and functionality of a brain. Deep learning models are neural networks, layers of interlinked individual unit cells, called neurons, connected to other neurons’ layers. Figure \ref{DANN} shows the deep feedforward neural network, consisting of linked layers of neurons that calculate an output layer (predictions) based on input data.

\FloatBarrier

Based on the input layer, the layers of neurons propagate information forward to the successive layers, creating a learning network with some feedback mechanism. A neural network’s depth is measured by the number of hidden layers, i.e., the layers between input and output layers. Neurons of consecutive layers are connected through accompanying weights $\boldsymbol{W}$ and biases $\boldsymbol{b}$. For a layer $l$, the output $\boldsymbol{\hat{Y}}^{l}$ is calculated as:
\begin{equation}\label{AEq1}
\begin{aligned}
    \boldsymbol{Z}^{l}&=\boldsymbol{W}^{l} \boldsymbol{\hat{Y}}^{l-1}+\boldsymbol{b}^{l}\\
    \boldsymbol{\hat{Y}}^{l}&=f^{l}\left(\boldsymbol{Z}^{l}\right)\\
\end{aligned}
\end{equation}
where the weights $\boldsymbol{W}$ and biases $\boldsymbol{b}$ are updated after every training pass. The activation function $f^{l}$ is an $\mathbb{R} \rightarrow \mathbb{R}$ mapping that transforms vector $\boldsymbol{Z}^{l}$, calculated using weights and biases, into output for every neuron in the layer $l$. In neural networks, the activation functions are nonlinear functions such as sigmoid, rectified linear unit (ReLu), and hyperbolic tangent. They empower the neural network to learn nearly any multifaceted functional correlation between inputs and outputs. After each feedforward pass, the loss function $\mathcal{L}$, such as the mean square error (MSE), calculates a loss value that indicates how well the network’s predictions compare with targets.

A loss function $\mathcal{L}$ is defined, and then it is used to obtain the weights $\boldsymbol{W}$ and biases $\boldsymbol{b}$ yielding a minimized loss value. The process of determining the optimized weights and biases in the context of machine learning is called training. Backpropagation is utilized throughout the training process, wherein the loss function is minimized iteratively. One of the most prevalent and most straightforward optimizers used is gradient descent \cite{pattanayak2017pro}:
\begin{equation}\label{AEq2}
\begin{aligned}
    W_{ij}^{c+1}&=W_{ij}^{c}-\gamma \frac{\partial \, \mathcal{L}}{\partial W_{ij}^{c}}\\
        b_{i}^{c+1}&=b_{i}^{c}-\gamma \frac{\partial \, \mathcal{L}}{\partial b_{i}^{c}}\\
    \end{aligned}
\end{equation}

where $\gamma$ represents the learning rate. Equation \ref{AEq2} shows the formula used to update the weights $\boldsymbol{W}$ and biases $\boldsymbol{b}$ at a given iteration $c$ within the gradient descent training process.

\subsection{Deep energy method}\label{DEM}

\begin{figure}[!htb]
    \centering
    \includegraphics[width=0.8\textwidth]{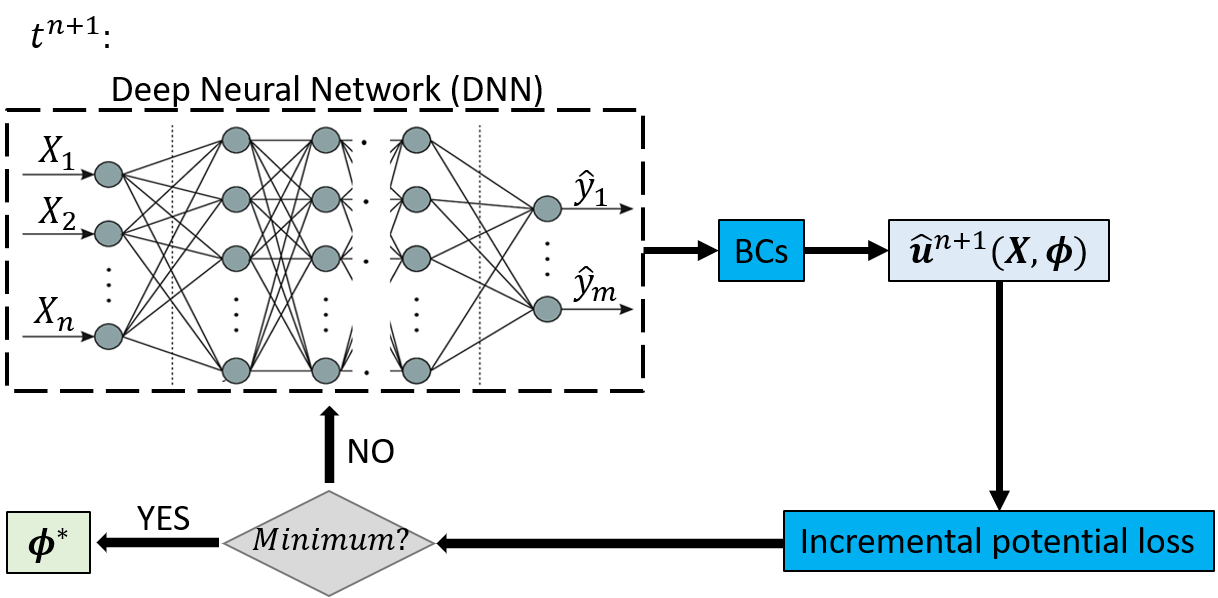}
    \caption{Flow chart of the deep energy method (DEM) at time increment $t^{n+1}$.}
    \label{DEM_FC}
\end{figure}

In this section, we discuss the deep energy method (DEM) in a general setting. The DEM uses an incremental potential that defines a loss function, which is minimized with the assistance of deep learning, as depicted in Figure \ref{DEM_FC}. The potentials corresponding to the material models are given in the following sections.

We consider a generalized partial differential equation (PDE), with solution $\boldsymbol{u}\left(t,\boldsymbol{X}\right)$, expressed as: 
\begin{equation}\label{PDEs}
\begin{aligned}
    \left(\partial_t + \mathcal{N}\right)\boldsymbol{u}\left(t,\boldsymbol{X}\right)&=\boldsymbol{0}{,} \quad  \left(t,\boldsymbol{X}\right) \in \left[0,T\right]\times\Omega,\\
    \boldsymbol{u}\left(0,\boldsymbol{X}\right)&=\boldsymbol{u}_o \quad  \boldsymbol{X} \in \Omega,\\
    \boldsymbol{u}\left(t,\boldsymbol{X}\right)&=\overline{\boldsymbol{u}}, \quad \! \left(t,\boldsymbol{X}\right) \in \left[0,T\right]\times\Gamma_{u},\\
    \boldsymbol{t}\left(t,\boldsymbol{X}\right)&=\overline{\boldsymbol{t}}, \quad  \left(t,\boldsymbol{X}\right) \in \left[0,T\right]\times\Gamma_{t},\\
\end{aligned}
\end{equation}
where $T$ is the total time, $\partial_t$ is the partial derivative with respect to time $t$, $\mathcal{N}$ is a spatial differential operator, $\overline{\boldsymbol{u}}$ is a defined essential boundary condition, $\boldsymbol{u}_o$ is the initial condition, and $\overline{\boldsymbol{t}}$ is a defined natural boundary condition. $\Omega$ is the material domain, while $\Gamma_t$ and $\Gamma_u$ are the surfaces with natural and essential boundary conditions, respectively. Here, we attempt to solve partial differential equations by training a neural network with parameters $\boldsymbol{\phi}= \{\boldsymbol{W}, \boldsymbol{b}\}$. Specifically, the model is trained such that the approximate solution $\boldsymbol{\hat{u}} \left(t,\boldsymbol{X}; \boldsymbol{\phi} \right)$ obtained from the neural network minimizes the incremental potential. The examples discussed in this paper assume a quasi-static condition, i.e., the term $\partial_{t} \boldsymbol{u} = \boldsymbol{0}$ and there is no need for defining initial conditions. However, the solution $\boldsymbol{u}^{n+1} \left( \boldsymbol{X} \right)$ is obtained incrementally.

Figure \ref{DEM_FC} depicts the deep energy method used here. We cast the minimization of the loss function $\mathcal{L}$ into the optimization problem in the context of deep learning, which is commonly called training. In other words, the DNN minimizes the loss function to obtain the optimized network parameters $\boldsymbol{\phi}^{*} =\{\boldsymbol{W}^*, \boldsymbol{b}^*\}$, where the loss function is defined by the incremental potential for a specific problem, and the neural network is used as global shape function for the displacement over the physical domain of interest. Specifically, the DNN maps the coordinates $\boldsymbol{X}$ of the sampled points to an output $\boldsymbol{\hat{y}}\left(\boldsymbol{X}, \boldsymbol{\phi}\right)$ using the feedforward propagation. Then, there are two approaches to satisfy the essential boundary conditions. The first approach is to subject $\boldsymbol{\hat{y}}\left(\boldsymbol{X}, \boldsymbol{\phi}\right)$ to:
\begin{equation}\label{NN_BC1}
\begin{aligned}
    \boldsymbol{\hat{u}}\left(\boldsymbol{X}, \boldsymbol{\phi}\right) &= \boldsymbol{A}\left(\boldsymbol{X}\right) + \boldsymbol{B}\left(\boldsymbol{X}\right) \circ \boldsymbol{\hat{y}}\left(\boldsymbol{X}, \boldsymbol{\phi}\right)\\
\end{aligned}
\end{equation}
where $\boldsymbol{A}\left(\boldsymbol{X}\right)$ and $\boldsymbol{B}\left(\boldsymbol{X}\right)$ are chosen such that the displacement field $\boldsymbol{\hat{u}}\left(\boldsymbol{X}, \boldsymbol{\phi}\right)$ satisfies the essential boundary conditions active on the physical domain. More details can be found in the work of Nguyen-Thanh et al. \cite{nguyen2020deep} and Rao et al. \cite{rao2021physics}. The second approach to satisfy the essential boundary conditions is to set:
\begin{equation}\label{NN_BC2}
\begin{aligned}
    \boldsymbol{\hat{u}}\left(\boldsymbol{X}, \boldsymbol{\phi}\right) &=  \boldsymbol{\hat{y}}\left(\boldsymbol{X}, \boldsymbol{\phi}\right)\\
\end{aligned}
\end{equation}
where one has to account for the boundary conditions throughout the training process of the DNN by adding an extra term to the loss function, as in the work of Abueidda et al. \cite{abueidda2020meshless}. In this paper the former approach is used since the latter is a soft enforcement of the boundary conditions and does not guarantee that the boundary conditions are imposed because of the pathology issue of the gradient. The predicted displacement field $\boldsymbol{\hat{u}}$ is then used to calculate the dependent variables and loss function. The computation of the dependent variables and loss function typically requires determining the first derivative of $\boldsymbol{\hat{u}}$, which is found using the automatic differentiation offered by deep learning frameworks. The optimization (minimization) problem is written as:
\begin{equation}\label{Minimization}
\begin{aligned}
    \boldsymbol{\phi}^{*}&=\underset{\boldsymbol{\phi}}{\mathrm{arg\,min}} \; \mathcal{L}\left(\boldsymbol{\phi}\right){.}\\
\end{aligned}
\end{equation}

\subsection{DNN model}\label{archit}

We use an $8$-layer neural network (depth$=6$), where the numbers of neurons in the successive layers are $3-40-40-40-40-40-40-3$. We  use three neurons in the input layer to handle 3D spatial coordinates $\boldsymbol{X}$, while we use three neurons in the output layer representing the components of the displacement vector $\boldsymbol{\hat{u}}\left(\boldsymbol{X}\right)$. The number of neurons utilized in each of the six hidden layers is $nHL=40$. An activation function was used after each layer, as demonstrated in Equation \ref{AEq1}. The architecture of the network is obtained by trying several architectures and fine-tuning of hyperparameters.

Abandoning nonlinear activation functions reduces the network to a linear one, making it challenging to capture nonlinear relationships between input and output. For example, some popular activation functions, in many machine learning practices, are ReLU and leaky ReLU. This paper uses a ReLU activation function in the input and hidden layers, while we use a linear activation function in the output layer. 

In this paper, we use PyTorch to solve the deep learning problem (minimize the loss function). The two optimizers are employed in a serial fashion; the Adam optimizer is initially used, and then the limited-memory BFGS (L-BFGS) optimizer with the Strong Wolfe line search method \cite{liu1989limited, lewis2013nonsmooth} is utilized. We found that using both optimizers regularly stabilizes the optimization procedure; a similar conclusion was sketched in other papers in the literature \cite{abueidda2020meshless, guo2020analysis}. More details are discussed in the following sections.

\section{Hyperelasticity}\label{hyper}

We consider a body made of a homogeneous and isotropic hyperelastic material under finite deformation. The mapping $\boldsymbol{\zeta}$ of material points from the initial configuration to the current configuration is determined by:
\begin{equation}\label{mapping}
\begin{aligned}
    \boldsymbol{x}&=\boldsymbol{\zeta}\left(\boldsymbol{X},t\right)=\boldsymbol{X}+\boldsymbol{\hat{u}}.\\
\end{aligned}
\end{equation}
Assuming that body and inertial forces are absent, the strong form is expressed as:
\begin{equation}\label{eqlbrm_HE}
\begin{split}
    \boldsymbol{\nabla}_{\boldsymbol{X}}\cdot \boldsymbol{P}&=\boldsymbol{0}{,} \quad  \boldsymbol{X}\in\Omega,\\
    \boldsymbol{\hat{u}}&=\overline{\boldsymbol{u}}, \quad \! \boldsymbol{X}\in\Gamma_{u},\\
    \boldsymbol{P}\cdot\boldsymbol{N}&=\overline{\boldsymbol{t}}, \quad  \boldsymbol{X}\in\Gamma_{t},\\
\end{split}
\end{equation}
where $\boldsymbol{\nabla}_{\boldsymbol{X}} \cdot$ is the divergence operator, $\boldsymbol{P}$ is the first Piola-Kirchhoff stress, and $\boldsymbol{N}$ represents the outward normal unit vector in the initial configuration. $\overline{\boldsymbol{u}}$ accounts for a defined essential boundary condition, and $\overline{\boldsymbol{t}}$ presents a defined natural boundary condition. $\Omega$ denotes the material domain, while $\Gamma_u$ and $\Gamma_t$ are the surfaces with essential and natural boundary conditions, respectively. The constitutive law for such a material is expressed as: 
\begin{equation}\label{constit_hyper}
\begin{aligned}
    \boldsymbol{P}&=\frac{\partial \psi\left(\boldsymbol{F}\right)}{\partial \boldsymbol{F}}\\
    \boldsymbol{F}&=\boldsymbol{\nabla}_{\boldsymbol{X}} \boldsymbol{\zeta}\left(\boldsymbol{X}\right)\\
\end{aligned}
\end{equation}
where $\psi$ is the strain energy density of a specific material, and $\boldsymbol{F}$ denotes the deformation gradient.
The material model of interest was proposed by Lopez-Pamies \cite{lopez2010new}. The strain energy model is given by:
\begin{equation}\label{strain_energy_hyper}
\begin{aligned}
    \psi &= \sum_{r=1}^{M} \frac{3^{1-\alpha_r}}{2 \alpha_r} \mu_r \left(I_1^{\alpha_r} - 3^{\alpha_r} \right) - \sum_{r=1}^{M} \mu_{r} \text{ln} J + \frac{\lambda}{2} \left(J-1\right)^{2}\\
\end{aligned}
\end{equation}
where $I_1$ is the the first principal invariant of the right Cauchy–Green deformation tensor $\boldsymbol{C}$, i.e., $I_1 = \text{trace}\left(\boldsymbol{C}\right)$. The right Cauchy–Green deformation tensor $\boldsymbol{C}$ is expressed as $\boldsymbol{C} = \boldsymbol{F}^T \boldsymbol{F}$, where $\left(\right)^{T}$ is the transpose operator. $M$ represents the number of terms included in the summation, while $\alpha_r$, $\mu_r$ and $\lambda$ are material constants $\left( r=1,2,..., M\right)$. $J$ is the determinant of the deformation gradient. We aim here at developing a DNN model that (1) is straightforward and amenable to numerical and analytical solutions for boundary-value and homogenization problems, (2) contains material constants providing a physical interpretation, and (3) characterizes and accurately predicts the mechanical behavior of rubber-like elastic solids over the entire range of deformations \cite{lopez2010new}. The constitutive relation implied by the potential shown in Equation (\ref{strain_energy_hyper}) is expressed as:
\begin{equation}\label{stress_hyper}
\begin{aligned}
    \boldsymbol{P} &= \frac{\partial \psi\left(\boldsymbol{F}\right)}{\partial \boldsymbol{F}} = \sum_{r=1}^{M} 3^{1-\alpha_r}\mu_r I_1^{\alpha_r-1} \boldsymbol{F} - \sum_{r=1}^{M} \mu_{r} \boldsymbol{F}^{-T} + \lambda \left(J^{2}-J\right) \boldsymbol{F}^{-T}\\
\end{aligned}
\end{equation}
In this paper, the number of terms $M$ in Equation (\ref{strain_energy_hyper}) is $M=2$. Following the deep energy method, one needs no incremental tangent modulus since we do not solve a linear system of equations here, as in the case of classical finite element problems. Hence, we do not include it.

To solve a hyperelasticity problem, we transform the strong form (see Equation (\ref{eqlbrm_HE})) into the weak form. However, for a hyperelastic material, the weak form can be expressed as:
\begin{equation}\label{hyper_weak}
\begin{aligned}
    \Pi = \underbrace{\int_{\Omega} \psi d\Omega}_{\text{internal energy}} \underbrace{- \int_{\Omega} \boldsymbol{u}^{T}\boldsymbol{f}_{b} d\Omega - \int_{\Gamma} \boldsymbol{u}^{T} \overline{\boldsymbol{t}} d\Gamma}_{\text{external energy}}
\end{aligned}
\end{equation}
For hyperelastic materials, we aim at minimizing the potential energy, shown in Equation (\ref{hyper_weak}), where the loss function to be minimized is defined as:
\begin{equation}\label{hyper_loss}
\begin{aligned}
    \mathcal{L} = \int_{\Omega} \hat{\psi}\left(\boldsymbol{X}, \boldsymbol{\phi}\right) d\Omega - \int_{\Omega} \boldsymbol{\hat{u}}^{T}\left(\boldsymbol{X}, \boldsymbol{\phi}\right) \;\boldsymbol{f}_{b} d\Omega - \int_{\Gamma} \boldsymbol{\hat{u}}^{T}\left(\boldsymbol{X}, \boldsymbol{\phi}\right)\; \overline{\boldsymbol{t}} d\Gamma\\
\end{aligned}
\end{equation}
where the body force $\boldsymbol{f}_{b}$ is assumed to be negligible in this work, and $\boldsymbol{\hat{u}}\left(\boldsymbol{X}, \boldsymbol{\phi}\right)$ is the approximate displacement obtained from the neural network, as shown in Figure \ref{DEM_FC}. A numerical integration scheme has to be used to calculate the loss function. In this work, we have used the 3D trapezoidal rule to evaluate the integrals defining the loss function shown in Equation (\ref{hyper_loss}). Algorithm \ref{algo_hyperelast} summarizes the steps involved in solving hyperelastic problems.  

\begin{algorithm}[!htb]
\SetAlgoLined
    \textbf{Input}: Physical domain, BCs, and DNN\\
    \quad \quad Material parameters ($\lambda, \mu_{1}, \mu_{2}, \alpha_{1}, \alpha_{2}$)\\ 
    \quad \quad Sample points $\boldsymbol{X}_{int}$ from $\Omega$\\
    \quad \quad Sample points $\boldsymbol{X}_{u}$ from $\Gamma_u$\\
    \quad \quad Sample points $\boldsymbol{X}_{t}$ from $\Gamma_t$\\
    \quad \quad Neural network architecture\\ 
    \quad \quad Neural network hyperparameters\\
    \quad \quad Optimizer (Adam followed by L-BFGS)\\
    \textbf{Initialization}: Initial weights and biases of the DNN\\
    \textbf{Output}: Optimized weights and biases of the DNN\\
    
 \While{Not minimized}{
    Obtain $\boldsymbol{\hat{u}}$ from the DNN\\
    Compute $\boldsymbol{\nabla}_{\boldsymbol{X}} \boldsymbol{\hat{u}}$ using automatic differentiation\\
    Compute $\boldsymbol{F}=\boldsymbol{I}+\boldsymbol{\nabla}_{\boldsymbol{X}} \boldsymbol{\hat{u}}$\\
    Compute $J=\text{det}\left(\boldsymbol{F}\right)$, $\boldsymbol{C}$, $I_1$, and $\boldsymbol{P}$\\
    Calculate loss function\\
    Update the weights and biases\\
 }
\caption{Hyperelasticity pseudocode.}
\label{algo_hyperelast}
\end{algorithm}

For illustrational purposes, we consider two scenarios: (1) a cube with unit length subject to a uniaxial tensile loading and (2) a cube with unit length subject to a simple shear loading. The material properties used are as follows: $\alpha_{1} = 1.0$, $\alpha_{2} = -2.47$, $\mu_{1}=13.5$ kPa, $\mu_{2}=1.08$ kPa, and $\lambda = 146.2$ kPa. 

\begin{figure}[!htb]
    \centering
    \includegraphics[width=1.0\textwidth]{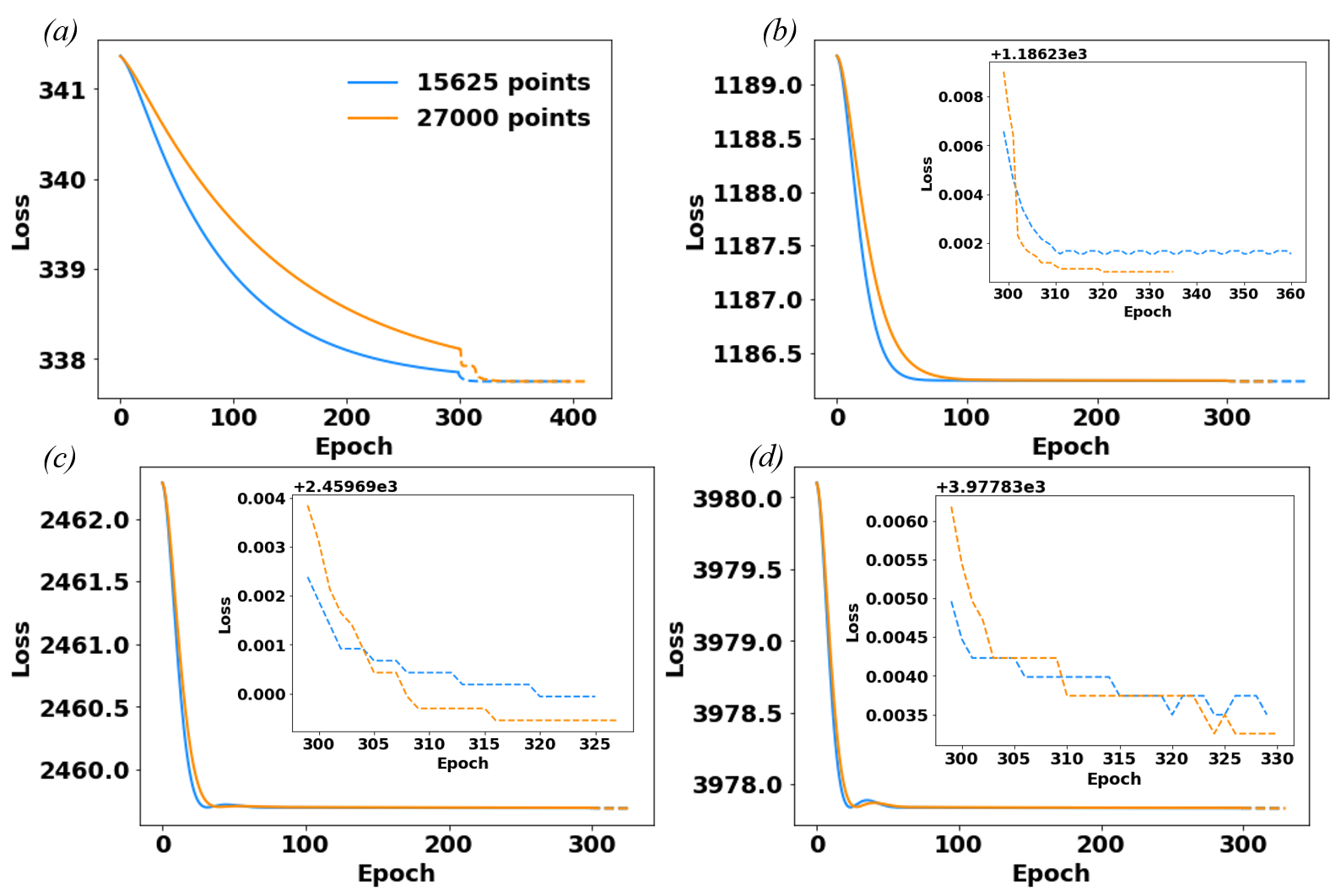}
    \caption{Hyperelasticity example (unit cube): Loss function convergence for different problem sizes (number of points). The solid lines represent the loss history obtained using the Adam optimizer, while the dashed lines portray loss history obtained from the L-BFGS optimizer. Different uniaxial strains are imposed: (a) $0.125$, (b) $0.25$, (c) $0.375$, and (d) $0.5$.}
    \label{Uniaxial_NPs}
\end{figure}

Here, we use a displacement-controlled approach. Considering a unit cube incrementally subject to a uniaxial strain of 0.5, the effect of the number of points, evenly spaced in the physical domain, on the convergence of the loss function is studied. The main reason for using evenly spaced points rather than randomly sampled points is to make the numerical integrations required to evaluate the loss function more straightforward. Figure \ref{Uniaxial_NPs} shows the convergence of the loss function. We have considered two problem sizes: $25\times25\times25 = 15625$ points and $30\times30\times30 = 27000$. Generally, we do not see a significant change in the converged values of the loss function when different numbers of points are taken from the physical domain. As discussed earlier, we have used two optimizers: the Adam optimizer followed by L-BFGS optimizer. For the Adam optimizer, we have used a fixed number, $300$ epochs, while for the L-BFGS optimizer, a tolerance of $1e-12$ was used as a stopping criterion. The solution is attained for $N$ displacement increments. Hence, the optimization problem is solved $N$ times. The optimized weights and biases found at increment $t$ are utilized to initialize weights and biases for the next increment $t+1$. This approach can be viewed as a form of transfer learning in the context and terminologies of machine learning, which is equivalent to updating displacements after each converged increment in any classical nonlinear implicit finite element analysis solution procedure. Due to the use of transfer learning, we observe that the convergence at later increments is faster and easier than those at the beginning, unlike the finite element method. Generally, we do not see a significant change in the convergence of the neural network discussed in section \ref{archit} when different numbers of points are taken from the physical domain.

\begin{figure}[!htb]
    \centering
    \includegraphics[width=1.0\textwidth]{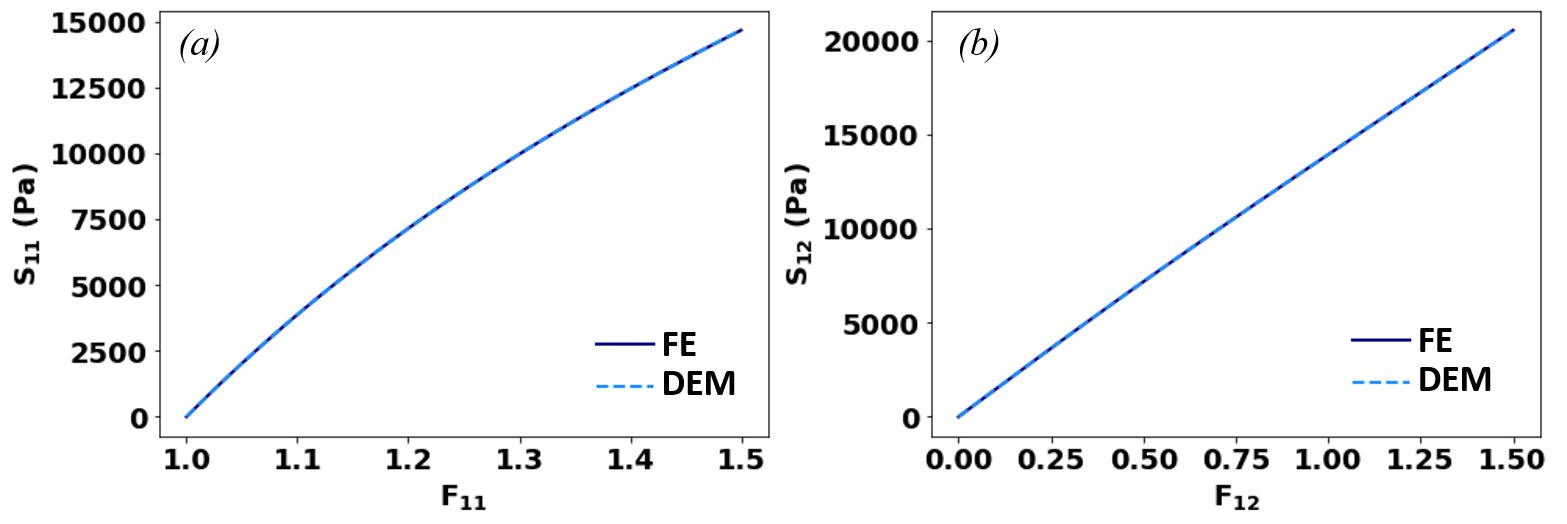}
    \caption{Comparison between the DEM and FEM results: Unit cube Subject to (a) uniaxial displacement and (b) simple shear.}
    \label{Hyper_results}
\end{figure}

Figure \ref{Hyper_results}a depicts a comparison between the solutions obtained from the traditional finite element analysis and the DEM. The DEM manages to capture the stress-deformation curve qualitatively and quantitatively. A similar analysis was done for the simple shear case, and similar conclusions were deduced. For the sake of brevity, we just include the final results, as shown in Figure \ref{Hyper_results}b. Since the hyperelasticity problem is path-independent, the DEM method does not require incremental loading. However, in classical finite element analysis, the problem is still solved incrementally in many cases to avoid divergence issues which we do not see in the  DEM method. Hence, if one is not interested in the intermediate evaluations, the DEM can solve the problem in one increment without having a pseudo time loop, as summarized in Algorithm \ref{algo_hyperelast}. However, in this paper, we are interested in the entire curve (see Figure \ref{Hyper_results}), which we have solved incrementally using transfer learning as discussed earlier.

\section{Viscoelasticity}\label{visco}

\begin{figure}[!htb]
    \centering
    \includegraphics[width=0.5\textwidth]{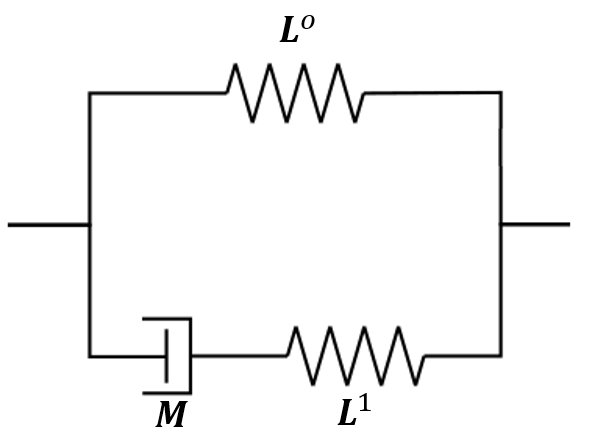}
    \caption{Rheological model of linear viscoelastic behavior.}
    \label{SLS_Model}
\end{figure}

We consider a 3D homogeneous, isotropic, linear viscoelastic model consistent with the standard linear solid (SLS) model depicted in Figure \ref{SLS_Model} under small deformation. The equilibrium equation, in the absence of inertial and body forces, is written as:

\begin{equation}\label{eqlbrm_E}
\begin{split}
    \boldsymbol{\nabla}\cdot \boldsymbol{\sigma}&=\boldsymbol{0}{,} \quad  \boldsymbol{x}\in\Omega,\\
    \boldsymbol{\hat{u}}&=\overline{\boldsymbol{u}}, \quad \! \boldsymbol{x}\in\Gamma_{u},\\
    \boldsymbol{\sigma}\cdot\boldsymbol{n}&=\overline{\boldsymbol{t}}, \quad  \boldsymbol{x}\in\Gamma_{t},\\
\end{split}
\end{equation}

where $\boldsymbol{\sigma}$ denotes the Cauchy stress tensor, $\boldsymbol{n}$ is the normal unit vector, $\boldsymbol{\nabla}\cdot$ is the divergence operator, and $\boldsymbol{\nabla}$ represents the gradient operator. Since small deformation is assumed, the strain tensor $\boldsymbol{\varepsilon}$ is expressed as:
\begin{equation}\label{eps}
\begin{aligned}
    \boldsymbol{\varepsilon}=\frac{1}{2}(\boldsymbol{\nabla} \boldsymbol{\hat{u}}+\boldsymbol{\nabla} \boldsymbol{\hat{u}}^T).\\
    \end{aligned}
\end{equation}
We adopt a two-potential constitutive framework to describe how a material stores and dissipates energy by defining two thermodynamic potentials: (1) a free energy function $\psi$ and (2) a dissipation potential $\eta$. The two potentials are defined by the equilibrium modulus of elasticity $\boldsymbol{L}^{o}$, the non-equilibrium modulus of elasticity $\boldsymbol{L}^{1}$, and the viscosity tensor $\boldsymbol{M}$. Assuming an isotropic material, the fourth-order tensors $\boldsymbol{L}^{o}$, $\boldsymbol{L}^{1}$, and $\boldsymbol{M}$ can be expressed as:
\begin{equation}\label{MaterialProp}
\begin{aligned}
    L^{o}_{ijkl} &= 2 \mu_{o} \mathcal{K}_{ijkl} + 3 \kappa_{o} \mathcal{J}_{ijkl}\\
    L^{1}_{ijkl} &= 2 \mu_{1} \mathcal{K}_{ijkl} + 3 \kappa_{1} \mathcal{J}_{ijkl}\\
    M_{ijkl} &= 2 \omega_{K} \mathcal{K}_{ijkl} + 3 \omega_{J} \mathcal{J}_{ijkl}\\
\end{aligned}
\end{equation}
with $\mathcal{K}_{ijkl}$ and $\mathcal{J}_{ijkl}$ defined as:
\begin{equation}\label{KJtensors}
\begin{aligned}
    \mathcal{K}_{ijkl} &= \frac{1}{2}\left(\delta_{ik}\delta_{jl} + \delta_{il} \delta_{jk} - \frac{2}{3} \delta_{ij}\delta_{kl}\right)\\
    \mathcal{J}_{ijkl} &= \frac{1}{3} \delta_{ij} \delta_{kl}\\
\end{aligned}
\end{equation}
where $\delta_{ij}$ denotes the Kronecker delta, $\mu$ is the shear modulus, $\kappa$ represents the bulk modulus, $\omega_{K}$ is the distortional viscosity, and $\omega_{J}$ is the volumetric viscosity.

The two thermodynamic potentials used to define the viscoelastic model are:
\begin{equation}\label{potentials}
\begin{aligned}
    \psi &= \frac{1}{2} \varepsilon_{ij} L^{o}_{ijkl} \varepsilon_{kl} + \frac{1}{2} \left(\varepsilon_{ij}-\varepsilon^{v}_{ij}\right) L^{1}_{ijkl} \left(\varepsilon_{kl}-\varepsilon^{v}_{kl}\right)\\
    &= \mu_{o} \varepsilon_{ij} \varepsilon_{ij} + \frac{3\kappa_{o}-2\mu_{o}}{6} \varepsilon_{ii} \varepsilon_{jj} + \mu_{1} \varepsilon^{e}_{ij} \varepsilon^{e}_{ij} + \frac{3\kappa_{1}-2\mu_{1}}{6} \varepsilon^{e}_{ii} \varepsilon^{e}_{jj}\\
    \eta &= \frac{1}{2} \dot{\varepsilon}^{v}_{ij} M_{ijkl} \dot{\varepsilon}^{v}_{kl}\\
    &= \omega_{K} \dot{\varepsilon}^{v}_{ij}\dot{\varepsilon}^{v}_{ij} + \frac{3\omega_{J}-2\omega_{K}}{6}\dot{\varepsilon}^{v}_{ii}\dot{\varepsilon}^{v}_{jj}\\ 
\end{aligned}
\end{equation}
where $\boldsymbol{\varepsilon}^{v}$ is the viscous part of the total strain $\boldsymbol{\varepsilon}$, and $\boldsymbol{\varepsilon}^{e} = \boldsymbol{\varepsilon} - \boldsymbol{\varepsilon}^{v}$ is the elastic part obtained from the additive decomposition. The constitutive relation implied by the above potentials is given by:
\begin{equation}\label{visco_stress}
\begin{aligned}
   \sigma_{ij} &= \frac{\partial\psi}{\partial\varepsilon_{ij}} = L^{o}_{ijkl} \varepsilon_{kl} + L^{1}_{ijkl} \left(\varepsilon_{kl} - \varepsilon^{v}_{kl}\right)\\
    &= 2\mu_{o} \varepsilon_{ij}+\frac{3\kappa_{o}-2\mu_{o}}{3}\varepsilon_{kk}\delta_{ij} + 2\mu_{1}\left(\varepsilon_{ij} - \varepsilon^{v}_{ij}\right) + \frac{3\kappa_{1}-2\mu_{1}}{3} \left(\varepsilon_{kk} - \varepsilon^{v}_{kk}\right) \delta_{ij} \\
\end{aligned}
\end{equation}
where the internal variable $\varepsilon^{v}_{kl}$ is implicitly defined by the evolution equation: 
\begin{equation}\label{EvolutionA}
\begin{aligned}
   \frac{\partial\psi}{\partial\varepsilon^{v}_{ij}} + \frac{\partial\eta}{\partial\dot{\varepsilon}^{v}_{ij}} &= 0\\
    -L^{1}_{ijkl} \left(\varepsilon_{kl} - \varepsilon^{v}_{kl}\right) + M_{ijkl} \dot{\varepsilon}^{v}_{kl} &= 0\\
\end{aligned}
\end{equation}
or equivalently:
\begin{equation}\label{EvolutionB}
\begin{aligned}
   \dot{\varepsilon}^{v}_{ij} &= M^{-1}_{ijmn} L^{1}_{mnkl} \left(\varepsilon_{kl} - \varepsilon^{v}_{kl}\right)\\
    &= \frac{\mu_{1}}{\omega_{K}}\left(\varepsilon_{ij} - \varepsilon^{v}_{ij}\right) + 
   \frac{1}{3}\left(\frac{\kappa_{1}}{\omega_{J}} - \frac{\mu_{1}}{\omega_{K}}\right)\left(\varepsilon_{kk} - \varepsilon^{v}_{kk}\right)\delta_{ij} \doteq G_{ij}\left(t,\boldsymbol{\varepsilon}^{v} \right)\\
\end{aligned}
\end{equation}
where the function $\boldsymbol{G}$, as a function of time $t$ and internal variable $\boldsymbol{\varepsilon}^{v}$, is defined for subsequent notational brevity.

The solution for the  problem of interest is obtained by defining the following incremental potential:
\begin{equation}\label{IncPotential}
\begin{aligned}
    \Pi &= \int_{\Omega} \psi\left(\boldsymbol{\varepsilon}, \boldsymbol{\varepsilon}^{v}\right) d\Omega + \Delta t \int_{\Omega} \eta\left(\dot{\boldsymbol{\varepsilon}}^{v}\right) d \Omega - \int_{\Omega} \boldsymbol{u}^{T}\boldsymbol{f}_{b} d\Omega - \int_{\Gamma} \boldsymbol{u}^{T} \overline{\boldsymbol{t}} d\Gamma\\
\end{aligned}
\end{equation}
where $\Delta t = t^{n+1} - t^{n}$. Assuming that the body forces $\boldsymbol{f}_{b}$ are zero and introducing the backward-Euler approximation:
\begin{equation}\label{Euler}
\begin{aligned}
    \dot{\boldsymbol{\varepsilon}}^{v} &\approx \frac{\boldsymbol{\varepsilon}^{v, n+1}-\boldsymbol{\varepsilon}^{v, n}}{\Delta t}\\
\end{aligned}
\end{equation}
into the dissipation potential, the incremental loss function is given by:
\begin{equation}\label{visco_loss}
\begin{aligned}
    \mathcal{L}^{n+1} &= \int_{\Omega} \psi\left(\boldsymbol{\varepsilon}^{n+1}, \boldsymbol{\varepsilon}^{v,n+1}\right) d\Omega + \Delta t \int_{\Omega} \eta\left(\frac{\boldsymbol{\varepsilon}^{v, n+1}-\boldsymbol{\varepsilon}^{v, n}}{\Delta t}\right) d \Omega  - \int_{\Gamma} {\boldsymbol{u}^{n+1}}^{T} \overline{\boldsymbol{t}}^{n+1} d\Gamma\\
\end{aligned}
\end{equation}
where $\boldsymbol{\varepsilon}^{v,n+1}$ is computed using the explicit fifth-order Runge–Kutta scheme with extended region of stability \cite{lawson1966order}, which is given by:
\begin{equation}\label{visco_loss}
\begin{aligned}
    \boldsymbol{\varepsilon}^{v,n+1} &= \boldsymbol{\varepsilon}^{v,n} + \frac{\Delta t}{90} \left(
    7 \boldsymbol{k}_1 + 32 \boldsymbol{k}_{3} + 12 \boldsymbol{k}_{4} + 32 \boldsymbol{k}_{5} + 7 \boldsymbol{k}_{6} 
    \right)\\
\end{aligned}
\end{equation}
where
\begin{equation}\label{RKScheme}
\begin{aligned}
    \boldsymbol{k}_1 &= \boldsymbol{G}\left(t_{n} , \boldsymbol{\varepsilon}^{v, n} \right)\\
    \boldsymbol{k}_2 &= \boldsymbol{G}\left(t_{n} + \frac{\Delta t}{2}, \boldsymbol{\varepsilon}^{v, n} + \boldsymbol{k}_1 \frac{\Delta t}{2}\right)\\
    \boldsymbol{k}_3 &= \boldsymbol{G}\left(t_{n} + \frac{\Delta t}{4}, \boldsymbol{\varepsilon}^{v, n} + \left(3\boldsymbol{k}_1 + \boldsymbol{k}_2\right) \frac{\Delta t}{16} \right)\\
    \boldsymbol{k}_4 &= \boldsymbol{G}\left(t_{n} + \frac{\Delta t}{2}, \boldsymbol{\varepsilon}^{v, n} + \boldsymbol{k}_3 \frac{\Delta t}{2}\right)\\
    \boldsymbol{k}_5 &= \boldsymbol{G}\left(t_{n} + \frac{3\Delta t}{2}, \boldsymbol{\varepsilon}^{v, n} + 3\left(-\boldsymbol{k}_2 + 2 \boldsymbol{k}_3 + 3 \boldsymbol{k}_4\right) \frac{\Delta t}{16}\right)\\
    \boldsymbol{k}_6 &= \boldsymbol{G}\left(t_{n} + \Delta t, \boldsymbol{\varepsilon}^{v, n} + \left(\boldsymbol{k}_1 +4 \boldsymbol{k}_2 +6 \boldsymbol{k}_3 -12\boldsymbol{k}_4+8\boldsymbol{k}_5\right) \frac{\Delta t}{7}\right).\\
\end{aligned}
\end{equation}

The procedure used to solve the viscoelastic problem is described in Algorithm \ref{algo_visco}. For viscoelastic problems, the solution is obtained using $M$ time steps. Hence, the optimization problem is solved $M$ times, where we use transfer learning such that the converged weights and biases obtained at the increment $t^{n}$ are used to initialize the DNN at the following increment $t^{n+1}$.

\begin{algorithm}[!htb]
\SetAlgoLined
    \textbf{Input}: Physical domain, BCs, and DNN\\
    \quad \quad Material parameters ($\kappa_{o}$, $\kappa_{1}$, $\mu_{o}$, $\mu_{1}$, $\omega_{K}$, and $\omega_{J}$)\\ 
    \quad \quad Sample points $\boldsymbol{X}_{int}$ from $\Omega$\\
    \quad \quad Sample points $\boldsymbol{X}_{u}$ from $\Gamma_u$\\
    \quad \quad Sample points $\boldsymbol{X}_{t}$ from $\Gamma_t$\\
    \quad \quad Neural network architecture\\ 
    \quad \quad Neural network hyperparameters\\
    \quad \quad Optimizer (Adam followed by L-BFGS)\\
    \quad \quad Initialize $\boldsymbol{\varepsilon}^{v}$\\
    \textbf{Initialization}: Initial weights and biases of the DNN\\
    \textbf{Output}: Optimized weights and biases of the DNN\\
 
\For{$t\gets1$ \KwTo number of steps}{
    Use weights and biases from previous step\\
     \While{Not minimized}{
        Obtain $\boldsymbol{\hat{u}}$ from the DNN\\
        Compute $\boldsymbol{\nabla} \boldsymbol{\hat{u}}$ using automatic differentiation\\
        Compute $\boldsymbol{\varepsilon}^{v}$ using the evolution equation\\
        Determine $\psi$ and $\eta$\\
        Calculate loss function\\
        Update the weights and biases\\
     }
}     
\caption{Linear viscoelasticity pseudocode.}
\label{algo_visco}
\end{algorithm}

\begin{figure}[!htb]
    \centering
    \includegraphics[width=1.0\textwidth]{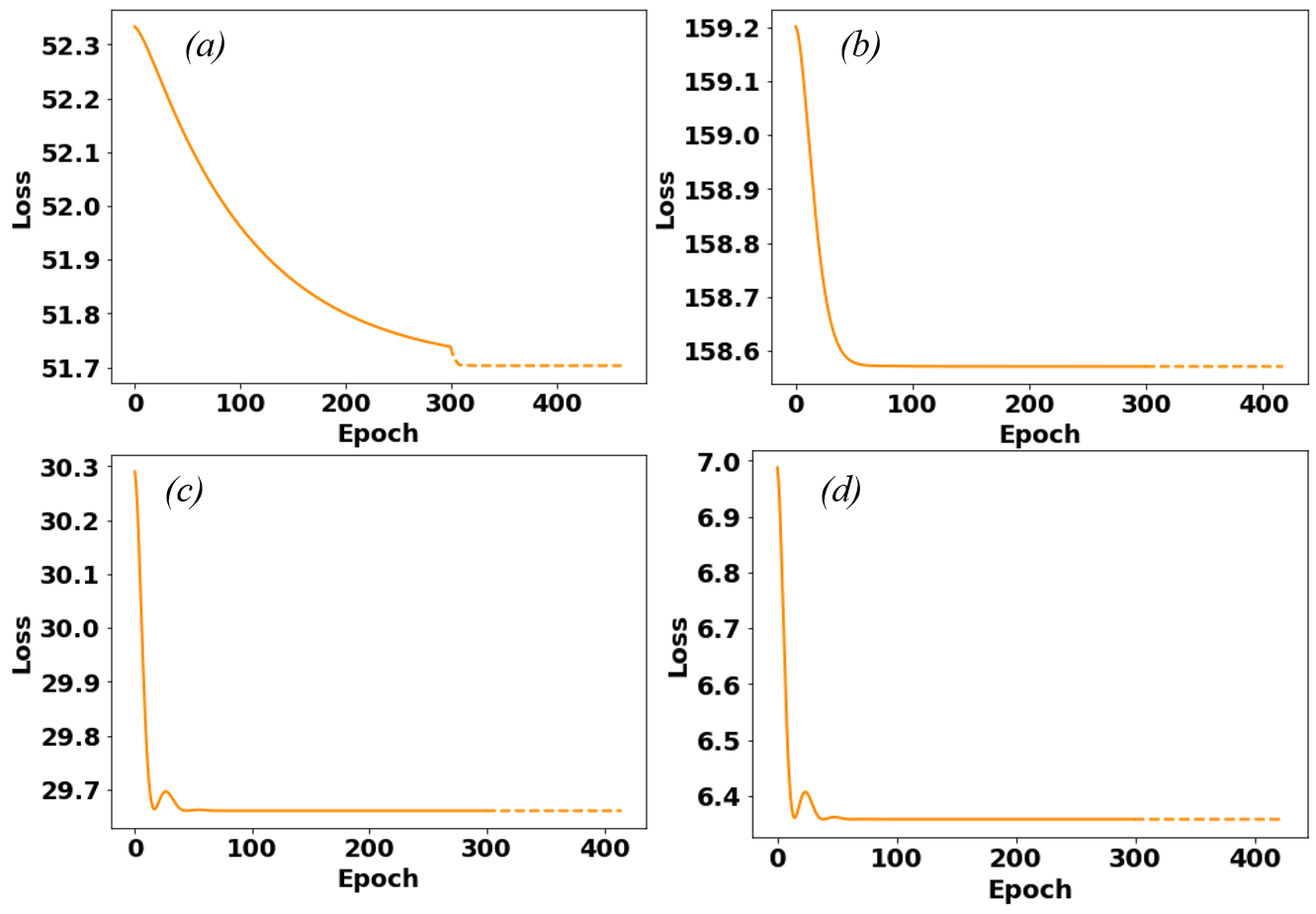}
    \caption{Two-potential viscoelasticity example (unit cube): Loss function convergence. The solid lines represent the loss history obtained using the Adam optimizer, while the dashed lines portray loss history obtained from the L-BFGS optimizer. Firstly, the cube is subject to a uniaxial strain of: (a) $1.5\%$ (at $t=0.5$ s) and (b) $3\%$ (at $t=1.0$ s). Then, it has been incrementally unloaded to a strain of (c) $1.5\%$ (at $t=1.5$ s) and (d) $0\%$ (at $t=2.0$ s).}
    \label{Load_Unload_Loss}
\end{figure}

\begin{figure}[!htb]
    \centering
    \includegraphics[width=1.0\textwidth]{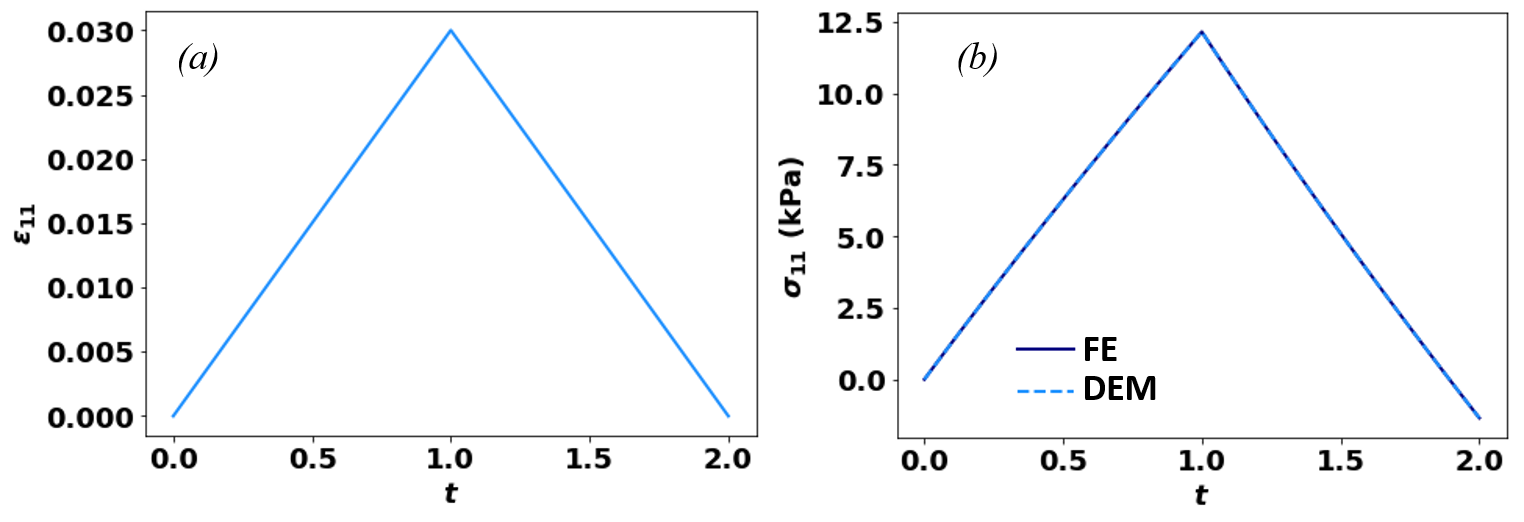}
    \caption{Loading and unloading: (a) History of the applied strain and (b) comparison between the results obtained from the FE analysis and DEM.}
    \label{Load_Unload}
\end{figure}

\begin{figure}[!htb]
    \centering
    \includegraphics[width=1.0\textwidth]{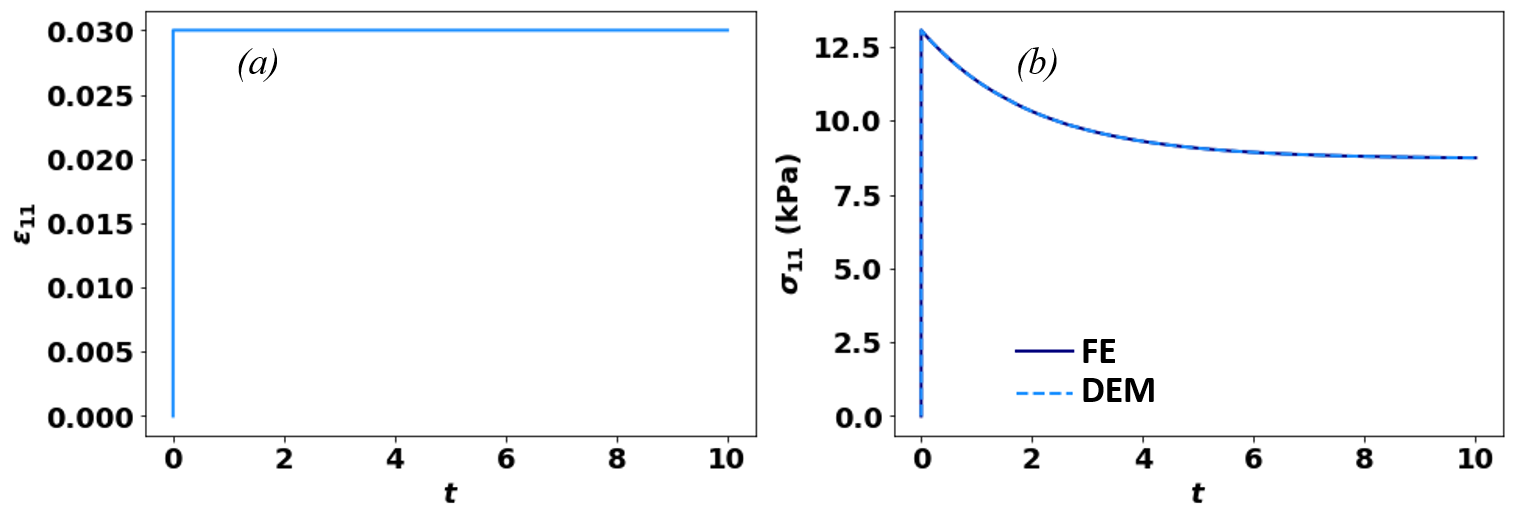}
    \caption{Relaxation test: (a) History of the applied strain and (b) comparison between the stress obtained from the FE analysis and DEM.}
    \label{Relaxation}
\end{figure}

Let's consider a unit cube under a loading and  unloading  uniaxial test. Specifically, it is incrementally subjected to a strain of 3\% in 1 s, and then it has been unloaded to a strain of 0\% in another 1 s. Figure \ref{Load_Unload_Loss} shows the convergence of the loss function at different strains. Figure \ref{Load_Unload} presents the history of the applied strain as well as the stress history. Another example we consider in this paper is the relaxation test. Figure \ref{Relaxation} depicts the applied strain and the relaxation of stress. The finite element results and those obtained from the DEM are in agreement in both test cases. 

\section{Discussion, conclusions, and future directions}\label{conclu}

In this study, the potential energy and deep learning are coupled to solve partial differential equations governing the mechanical deformation of hyperelastic and viscoelastic materials. Specifically, the first pillar of the DEM is that the potential energy that describes the mechanical deformation in a physical domain constitutes the basis for the construction of the loss function. The second pillar of the DEM is that the DNN defines the approximation space. When we solve for a hyperelastic response, the potential energy is minimized using deep learning tools. However, in the case of a viscoelastic response, an incremental potential, based on the two-potential approach, has to be defined and minimized. It is worth noting that this is one of the first implementations of 3D nonlinear solid mechanics using physics-informed neural networks. Using this approach, no data is created, which is often the bottleneck stage in constructing a data-driven model, and instead, the deep energy method (DEM) is used to get the solution. Compared to the physics-informed solution based on the strong form, the DEM solution is generally smoother. Furthermore, the DEM is meshfree; hence, we do not define connectivity between the nodes and mesh generation, which can be challenging in many cases \cite{bourantas2018strong} and may necessitate partitioning methods for large meshes \cite{borrell2018parallel}. Also, meshfree methods do not encounter the issue of element distortion or volumetric locking.

As stated in Section \ref{NN}, the deep learning approach is based mainly on matrix-vector multiplications, which are highly well-tuned and execute on GPUs. Unfortunately, this is not the case with direct sparse solvers in the implicit finite element method, which are frequently the only robust solver technique, but to date, it fails to make full use of GPUs. For nonlinear high fidelity simulations with millions of degrees of freedom, it is realistic to predict that physics-informed neural networks using transfer learning on GPUs may outperform a traditional implicit finite element technique. Additionally, writing from scratch a physics-informed neural network to solve PDEs is much easier and quicker than writing a nonlinear finite element code. Finally, a combination of data-driven and physics-informed deep learning approaches has been used to solve incomplete and ill-posed problems in computational mechanics and other physics-based modeling disciplines, previously considered insolvable by the classical numerical methods \cite{cai2021flow}.

Other geometries, such as irregular 3D geometries, will be examined in future studies. High-gradient solution regions appear when the geometries get more complicated. A new extended deep energy formulation, known as the mixed deep energy method (mDEM) \cite{FUHG2021110839},  has been developed recently that predicts stresses in addition to displacements from the neural network and delivers more accurate findings in high-gradient solution portions of domains, such as stress concentrations. Another limitation of this work is that we picked evenly spaced points from the physical domain to make the numerical integrations required to evaluate the loss functions straightforward. However, in principle, one can adopt other random sampling techniques (such as Latin hypercube, Halton sequence, etc.) along with more sophisticated numerical integration schemes to investigate how different sampling techniques impact the model's performance and explore which ones lead to higher accuracy. Guo et al. \cite{guo2020analysis} studied the effect of sampling techniques in the context of the deep collocation method (DCM), and it would be interesting to examine this effect in the context of the DEM. 

The optimization problems solved using the DCM or DEM are often non-convex. Hence, one must be attentive to getting trapped in local minima. Nonconvexity leads to several challenges that have to be investigated by the mechanics community. Additionally, in this paper, the architecture of the network and hyperparameters are chosen on a trial and error basis. However, one needs an architecture and hyperparameters that provide good accuracy and requires the architecture and hyperparameters to be optimized to get the highest performance possible. Hamdia et al. \cite{hamdia2020efficient} suggested using the genetic algorithm to find the optimized architecture and hyperparameters. Solving PDEs with physics-informed deep learning techniques is currently a growing trend within the research community, and our paper is by no means the final word on the subject.
\section*{Acknowledgment}
The authors would like to thank the National Center for Supercomputing Applications (NCSA) Industry Program and the Center for Artificial Intelligence Innovation.
\section*{Data availability}
The data that support the findings of this study are available from the corresponding author upon reasonable request. 

\bibliography{mybibfile}

\end{document}